\pdfoutput=1

\documentclass[11pt]{article}

\usepackage[preprint]{acl}

\usepackage{times}
\usepackage{latexsym}

\usepackage[T1]{fontenc}

\usepackage[utf8]{inputenc}

\usepackage{microtype}

\usepackage{inconsolata}

\usepackage{graphicx}
\usepackage{booktabs,tabularx}
\usepackage{multirow}
\usepackage{enumitem}
\usepackage{amsmath}
\title{From Conversation to Automation: Leveraging LLMs for Problem-Solving Therapy Analysis}

\author{
  \makebox[\textwidth][c]{%
    Elham Aghakhani$^{1}$,
    Lu Wang$^{2}$,
    Karla T. Washington$^{3}$} \\
  \makebox[\textwidth][c]{%
    \textbf{George Demiris}$^{4}$,
    \textbf{Jina Huh-Yoo}$^{2}$,
    \textbf{Rezvaneh Rezapour}$^{1}$} \\
  \makebox[\textwidth][c]{%
    $^{1}$Drexel University, \texttt{\{ea664,sr3563\}@drexel.edu}} \\
  \makebox[\textwidth][c]{%
    $^{2}$Stevens Institute of Technology, \texttt{\{lwang97,jhuhyoo\}@stevens.edu}} \\
  \makebox[\textwidth][c]{%
    $^{3}$Washington University, \texttt{kwashington@wustl.edu}} \\
  \makebox[\textwidth][c]{%
    $^{4}$University of Pennsylvania, \texttt{gdemiris@nursing.upenn.edu}}
}

\begin{document}
\maketitle

\begin{abstract}
Problem-solving therapy (PST) is a structured psychological approach that helps individuals manage stress and resolve personal issues by guiding them through problem identification, solution brainstorming, decision-making, and outcome evaluation. As mental health care increasingly adopts technologies like chatbots and large language models (LLMs), it is important to thoroughly understand how each session of PST is conducted before attempting to automate it. 
We developed a comprehensive framework for PST annotation using established PST Core Strategies and a set of novel Facilitative Strategies to analyze a corpus of real-world therapy transcripts to determine \textit{which} strategies are most prevalent. Using various LLMs and transformer-based models, we found that GPT-4o outperformed all models, achieving the highest accuracy (0.76) in identifying all strategies.
To gain deeper insights, we examined \textit{how} strategies are applied by analyzing Therapeutic Dynamics (autonomy, self-disclosure, and metaphor), and linguistic patterns within our labeled data. 
Our research highlights LLMs' potential to automate therapy dialogue analysis, offering a scalable tool for mental health interventions. Our framework enhances PST by improving accessibility, effectiveness, and personalized support for therapists.
\end{abstract}

\section{Introduction}\label{sec:intro}
In recent years, modern stressors like COVID-19, global conflicts, and climate disasters have led to rising mental health conditions, especially among adults 35-44 \citep{apa2023}. In response, the World Health Organization (WHO) calls for urgent transformation in mental health care \citep{WHO2022}. Problem-Solving Therapy (PST) helps manage these stressors by enhancing problem-solving skills, reducing psychological distress, and strengthening coping mechanisms \citep{song2019efficacy, arean2008effectiveness, gellis2008problem}.

\begin{figure}
    \centering
    \includegraphics[width=.9\linewidth]{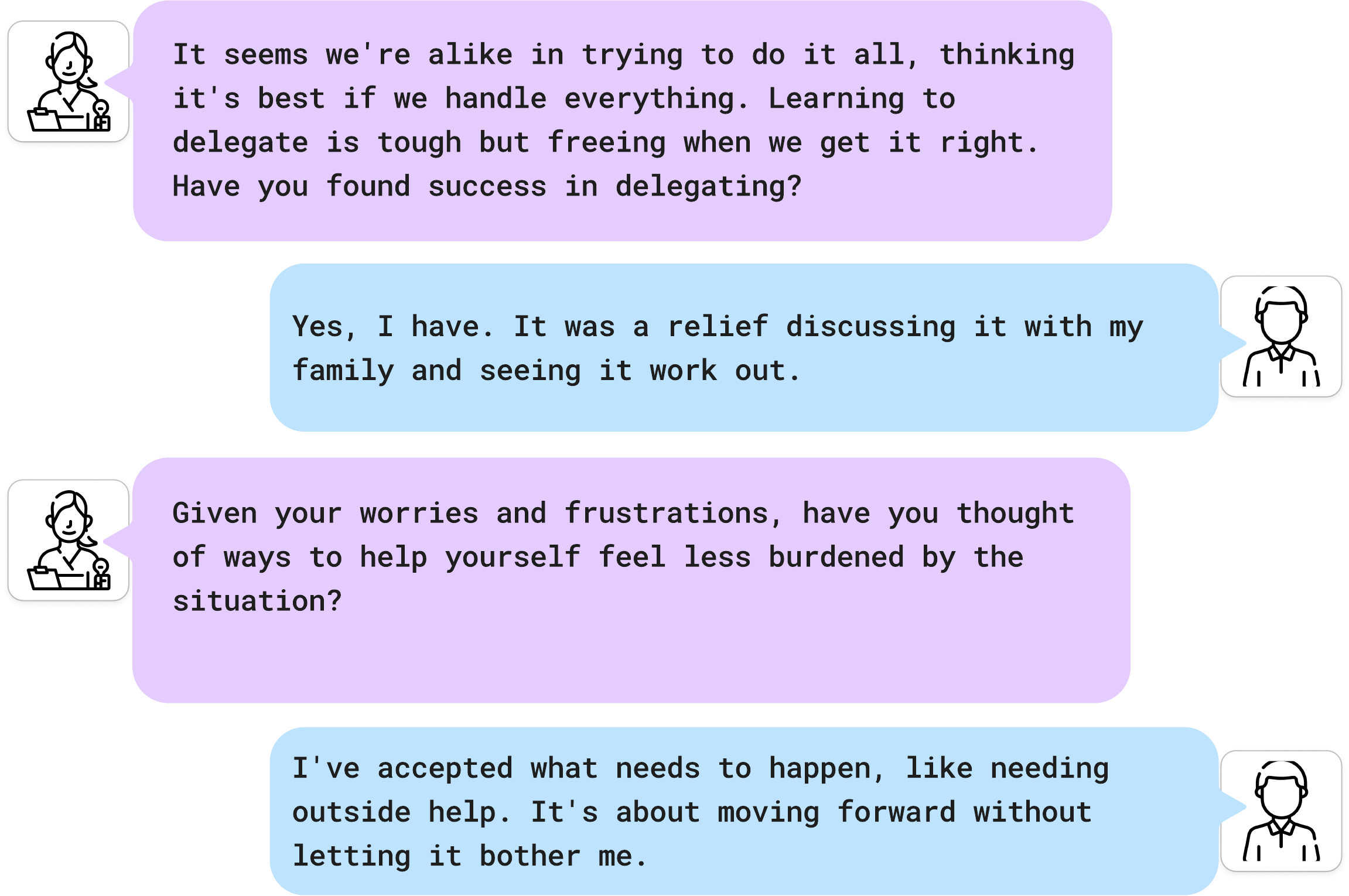}
    \caption{A real-life example of therapist and client exchanges in a problem-solving therapy session. }
    \label{fig:example}
    \vspace{-0.5cm}
\end{figure}

As rapidly evolving trends continue to shape the critical area of mental health, systematic methods for analyzing PST interactions remain limited \citep{chiu2024computational}. This gap is especially relevant with the rise of LLM-powered mental health chatbots \citep{lawrence2024opportunities, cho-etal-2023-integrative}. Before widespread adoption, it is crucial to assess how PST strategies are applied in practice, as therapist approaches vary based on client needs and contexts \citep{Andrade-Arenas_Yactayo-Arias_Pucuhuayla-Revatta_2024}. PST follows a structured process—identifying problems, brainstorming solutions, decision-making, and evaluation—to strengthen clients' problem-solving skills and coping abilities \citep{malouff2007efficacy}. While new technologies hold promise, their adaptability and responsiveness must be ensured for effective PST automation \citep{ai5020041}.


Previous work leveraged LLMs for analyzing therapist-client interactions \cite{lin2024compass,Demszky2023UsingLL}. For instance, \citeauthor{lee2024towards} used LLMs, particularly ChatGPT, to identify utterance-level features in therapist-client conversations to classify counselor's utterances. In PST, such research in automated annotations is still lacking due to limited data and costly domain expert annotation. Automating the annotation of various cognitive therapy techniques is crucial for developing efficient models, like LLMs, that can navigate the complexity of therapeutic dialogues and identify effective interventions. This capability could enable real-time suggestions to therapists or direct interventions in automated systems.

Our work addresses this gap by using LLMs to automate PST annotation, using both closed- and open-weight models to categorize therapist utterances. We analyzed 240 real-world anonymized PST sessions (68,306 dialogue exchanges), applying LLMs to annotate strategies based on existing and adapted PST and Facilitative strategy codebooks. Our results show that LLMs efficiently and accurately annotate PST strategies. We also analyzed how these strategies evolve across sessions finding that therapists progressively shift from exploratory strategies, such as defining problems and brainstorming solutions, to strategies focused on implementation and evaluation. Additionally, we analyze therapeutic dynamics (autonomy, self-disclosure, metaphor) and linguistic properties in all utterances to understand \textit{how} these strategies were applied. Our results show that therapists adapt their language to session goals—using more non-directive language and metaphors in early stages and transitioning to directive language and concrete communication as therapy progresses.

This approach enhances therapist-client interaction analysis, supports scalable and adaptive mental health interventions, and enables real-time identification of effective therapeutic techniques. Our framework advances NLP and mental health care, showing LLMs' potential to transform PST research. Finally, integrating Facilitative strategies refines the PST framework, offering deeper insights into targeted therapeutic interventions.






\section{Related Work} \label{sec:relatedwork}
\subsection{LLMs for Mental Health Care}
LLMs have been applied for mental health care, leveraging distinct features such as conversational capabilities \cite{ma2023understanding, zheng2023building}, content generation \cite{yang2024mentallama, smith2023old}, and detection of information patterns \cite{posner2011columbia, mauriello2021sad}. 
Specifically, LLMs have been applied by conversational agents~\citep{bendig2022next} to serve as personal digital companions \citep{ma2023understanding}, provide emotional support \citep{zheng2023building}, and support online counseling \citep{sharma2023cognitive, filienko2024toward}. 
Researchers applied LLMs to generate explanations for mental health analysis~\cite{yang2024mentallama}, help people learn mindfulness~\cite{kumar2023exploring}, and create hypothetical case vignettes to social psychiatry~\citep{smith2023old}. LLMs have also been used in classification of subtypes of suicide \citep{posner2011columbia}, and identification of sources of stress span \citep{mauriello2021sad}. 

However, gaps remain in previous research in accessing data with high external validity to fully understand the potential LLMs for different mental health care settings~\citep{hua2024large, wang2021evaluation, goel2023llms}. 
For example, among 36 datasets identified in \citet{hua2024large}, 55.6\% were collected from social media, 13.9\% were simulated by clinicians, crowd workers, and LLMs, and only 8.3\% came from real-life counseling conversations. Moreover, labels in existing datasets are typically not validated by experts~\citep{hua2024large, harrigian2020models}. The absence of relevant evaluation frameworks brings further challenges. 
Every cognitive and interpersonal therapy technique has a unique, specialized structure and requirement for evaluating fidelity of delivery\citep{sharma2023cognitive, mehta2022psychotherapy}. 
Examining how professionals deliver interventions through therapy techniques is essential for addressing these gaps. 

\subsection{LLMs as (Co-)Annotators}
Acquiring high-quality labeled data is time-consuming and costly when relying solely on human experts for annotation~\cite{wang2021want, wang2024human,bouzoubaa-etal-2024-decoding}. LLMs have been explored to support annotations through prompt engineering~\cite{reynolds2021prompt, goel2023llms,bouzoubaa-etal-2024-words}, such as zero-shot, without any in-domain training~\cite{liu2024logprompt, reynolds2021prompt} and few-shot learning, incorporating examples of the task~\cite{brown2020language}. Research showed LLMs outperformed crowd workers on annotation tasks including relevance, stance, topics, frame detection, and categorizing sentence
segments \cite{gilardi2023chatgpt, he2023annollm, tornberg2023chatgpt}. 

On the other hand, \citet{zhu2023can} and \citet{ziems2024can} showed that the performance of LLMs
vary depending on tasks, datasets, and labels, claiming the irreplaceableness of human annotators. \citet{wang2021want} and \citet{li2024comparative} found that human-LLM collaboration—either through human annotation of LLM-labeled data or combining both annotations—outperformed LLM-only or human-only approaches. \citet{li2023coannotating} explored the allocation between humans and LLMs for annotation and contributed to a paradigm using uncertainty to estimate LLMs' annotation capability. Researchers also developed human-LLM collaborative annotation systems to support the annotations~\cite{kim2024meganno+, tang2024pdfchatannotator}.
Our study expands on previous research by employing LLMs to annotate therapy transcripts, thereby refining existing frameworks and models and informing broader applications of LLMs in mental health care research.
\section{Method} \label{sec:method}

\subsection{Data}

\noindent\textbf{Data Collection. }
[Anonymized collaborator] provided transcriptions of 240 PST sessions between 152 family caregivers of older adults in senior living facilities and their therapists. Each caregiver attended up to three therapy sessions, but not every session was recorded and transcribed due to technical and legal constraints. We first ensured the anonymity of these transcripts by removing any remaining information that could potentially reveal the identity of the clients or therapists. 
We then converted the transcripts into 68,306 dialogue exchanges between clients and therapists. An example of two exchanges is shown in Figure \ref{fig:example}. We then separated the utterances\footnote{An utterance is a continuous piece of speech, by one person, before or after which there is silence on the part of the person: \url{https://en.wikipedia.org/wiki/Utterance}} made by therapists, which amounted to 34,273 utterances. After consulting with the domain experts and to ensure high quality and relevance, we excluded utterances shorter than 5 words, such as ``hmmm'', ``okay'', ``well'', and ``oh good'' assuming that longer ones would provide more meaningful content for our study. 
This process resulted in 14,417 utterances for further analysis. The average word count for these utterances was 29.78 with a standard deviation of 42.72.

\noindent\textbf{Problem-Solving Therapy Codebook.} In collaboration with PST experts,  we developed the annotation codebook using the core strategies from the ADAPT model of PST, which includes five steps: 1) developing a positive \textbf{A}ttitude, 2) \textbf{D}efining the problem and setting goals, 3) being creative and finding \textbf{A}lternative solutions, 4) \textbf{P}redicting how the solutions will work and making a plan, and 5) \textbf{T}rying out the solution plan~\cite{demiris2019problem}. We refer to these as the `Problem-Solving (PS) Core' category. Upon further exploration of our data and various frameworks and codebooks for therapy conversations and strategies~\cite{li2023understanding, ribeiro2013collaboration, wang2023tpe, ghamari2022tracking}, our domain experts suggested adding a new category of four strategies to this list, called `Facilitative' strategies, which we will use interchangeably with `Facilitators' throughout the paper. These strategies are further explained in Table \ref{codebook}. 

\noindent\textbf{Data Annotation.} To develop an evaluation set, we randomly selected 500 therapist utterances for detailed annotation. One domain expert alongside one researcher from our team performed the annotation. The annotators were instructed to select at most one strategy from the `PS Core' category and one strategy from the `Facilitators', if applicable. If none of the strategies were suitable, we labeled it as `None'. We used Cohen's Kappa to assess annotators' agreement across different classes of PST strategies. Our findings show that the agreement scores ranged from 0.69 to 0.88, indicating a substantial agreement among the annotators, reflecting a reasonable level of consistency in identifying the strategies. Agreement was higher for more straightforward classes (e.g., `Generating Alternative Solutions') and lower for more subjective or nuanced classes (e.g., `Therapeutic Engagement'). 
   
\subsection{Classification of PST Strategies}
\noindent\textbf{LLM Prompting. } LLMs, such as GPT-4, have been used for data annotation, offering capabilities for understanding and categorizing diverse datasets \citep{li2023understanding, kuzman2023chatgpt, hoes_altay_bermeo_2023}. In this work, we employed these models to annotate therapist utterances, aiming to identify the PST strategy used in each case. Specifically, we used two versions of OpenAI's models: \texttt{gpt-4o-2024-08-06} \citep{openai_gpt_4} (referred to as GPT-4o) and \texttt{gpt-4-0125-preview} \citep{gpt40125preview} (referred to as GPT-4), as well as two open-source models, \texttt{Llama-3.1-70B-Instruct} \citep{meta_llama_3_1} (referred to as Llama) and \texttt{Yi-1.5-34B} \citep{young2024yi} (referred to as Yi-1.5). All models were deployed with a zero-temperature setting to ensure deterministic output, enhancing the consistency of our annotations (See Appendix \ref{app:model} for hyperparameters). Using this approach, we compared the efficiency and accuracy of these models against our 500 annotated utterances.
We used two approaches to test how well LLMs can annotate data:

\begin{itemize}[noitemsep,topsep=0pt]
    \item[-] \textbf{No Context:} We provided specific instructions to LLMs, followed by definitions of each strategy and a few examples. We then presented each therapist's utterance for classification, without providing any additional conversational context.
    \item[-] \textbf{Two Previous Utterances as Context:} In addition to instructions, strategy definitions, and examples, we provided LLMs with two previous statements from the conversation—one from the client and one from the therapist. Using this context, we evaluated how LLMs perform when conversation history is added.
\end{itemize}

Labeling with LLMs followed the same approach as with the annotators. In each prompt, we asked the models to select one strategy from the PS Core and Facilitative categories or return `None' if applicable. Prompts are available in Appendix \ref{Appendix_prompts}.
We ran each LLM five times to test their stability for annotation. To evaluate the consistency of the annotation, we calculated the entropy of the labels generated by the LLMs \citep{li-etal-2023-coannotating}. Entropy (Eq. \eqref{eq:entropy}) measures the degree of uncertainty or disorder within a dataset. The higher the entropy value, the more uncertain the LLMs' responses are. 
\vspace{-0.25cm}
\begin{equation}
u_i = - \sum_{j=1}^{k} P_{\theta}(a_{ij} \mid p_{ij}) \ln P_{\theta}(a_{ij} \mid p_{ij})
\label{eq:entropy}
\end{equation}

where $P_{\theta}(a_{ij}\mid p_{ij})$ is calculated as the frequency of one prediction among all predictions.

\noindent\textbf{Transformer-Based Models. }In addition to using LLM-based models, we tested the performance of four transformer-based models—DeBERTa \cite{he2020deberta}, MentalBERT \cite{ji-etal-2022-mentalbert}, FlanT5 \cite{chung2024scaling}, and ModernBERT \cite{modernbert}—on classifying utterances. From the 14,417 utterances annotated by our best LLM, we randomly selected 5,000 utterances, excluding the evaluation set, and fine-tuned the models on this subset. The models were fine-tuned separately for the two annotation dimensions, PS Core and Facilitative categories, and evaluated on the 500-utterance human-annotated test set.

\subsection{Therapeutic Dynamics}
Previous section focused on identifying `which' strategies are used. Here, we turn our attention to `how' these strategies are applied by analyzing autonomy, self-disclosure, and metaphor usage. 

\noindent\textbf{Autonomy} is a cornerstone of effective counseling, enhancing therapeutic alliances and treatment outcomes \cite{dwyer2011participant}. While its importance is well-established, measuring autonomy has largely focused on theory and behavior, overlooking linguistic markers. Previous research \cite{10.1145/3555640} has often relied on open-ended versus close-ended questions as a proxy for assessing autonomy. However, we adopt a more comprehensive definition grounded in the literature to better capture the nuances of autonomy in therapeutic interactions. To address this, we define autonomy through three categories: directive, non-directive, \cite{overholser1987facilitating} and N/A. Directive approaches involve therapist-led guidance, useful but potentially fostering dependency. Non-directives, such as Socratic questioning, encourage independence and self-reliance. The N/A category covers utterances unrelated to autonomy. 

\noindent\textbf{Self-disclosure} in therapy involves a therapist sharing personal experiences or feelings to build trust \cite{marais2021therapists}, validate emotions, or model behaviors, enabling a more equitable power relationship in therapy and normalizing clients' experiences and distress \cite{jolley2019m}. It can be categorized as immediate, involving real-time feelings, or non-immediate, sharing past experiences and personal anecdotes \cite{farber2006self}. Our framework also includes a None category for utterances lacking self-disclosure.


\noindent\textbf{Metaphor} is a symbolic approach that implies similarity between experiences, thoughts, emotions, actions, or objects \cite{wagener2017metaphor}. Given their crucial role in fostering learning and comprehension, counselors must be highly skilled at using them effectively in therapy~\cite{evans2010figurative}. While some studies show that LLMs struggle with metaphor and analogy comprehension \cite{czinczoll-etal-2022-scientific, tong-etal-2024-metaphor}, we incorporated key concepts from the Conceptual Metaphor Theory \cite{tian-etal-2024-theory}, which has shown to enhance LLMs performance in metaphor understanding. We designed our prompts to identify metaphors, along with their corresponding source and target domains.
We prompted GPT-4o to analyze therapeutic dynamics in each therapist utterance (see Appendix \ref{Appendix_prompts_domain}).

\vspace{-0.2cm}
\subsection{Linguistic Patterns}
We used the Linguistic Inquiry and Word Count (LIWC) tool \cite{boyd2022development} to measure the frequency of words across various linguistic categories within the utterances.

\vspace{-0.2cm}
\section{Result} \label{sec:result}
\subsection{Performance of LLMs in Annotation}
Table \ref{tab:f1_scores} shows the weighted F1 scores of four LLMs, including two proprietary models (GPT-4 and GPT-4o) and two open-source models (Llama-3.1 and Yi-1.5). We evaluate the models' performance using our evaluation set (500 annotated utterances) for both `No context' and `With context' conditions.
GPT-4o achieved the highest F1 score (0.76) with `No context', outperforming all other models. 
GPT-4 followed closely with F1 score of 0.68 (See Appendix \ref{Appendix_llm_results} for more details).
All models showed lower F1 scores when provided with additional context (i.e., the two previous utterances in the conversation), and our error analysis revealed that this was because, despite explicitly instructing the LLMs in our prompt to focus only on the therapist's most recent utterance and use the two previous utterances solely as context, the LLMs occasionally returned labels for the context rather than the target utterance.

Table \ref{tab:entropy_comparison} shows entropy results across five runs per LLM. Since values were identical for `No context' and `With context', we report a single set, with all models showing sufficiently low entropy for consistency.
GPT-4 and GPT-4o had higher mean entropy than Llama-3.1 and Yi-1.5, with GPT-4o slightly more stable (0.035 vs. 0.042). Despite higher variability, GPT models' superior F1 scores suggest this had no impact on performance. In contrast, Llama-3.1 and Yi-1.5 exhibited lower entropy, with Yi-1.5 at 0.00, potentially indicating overconfidence and poor generalization, reflected in its lower F1 score.

Overall, the results show that GPT-4o performed best without any context. Precision, recall, and F1 scores for each PST strategy are shown in Table \ref{tab:gpt4o_scores}. The model performed well in classifying PS Core strategies, with balanced scores across strategies like `Problem-Solving Positive Mindset' and `Defining Problems and Goals'. For Facilitators, it excelled in identifying `Therapeutic Engagement' and `Problem-Solving Test Review', while `Social Courtesies' and `Session Management' showed slightly lower precision. 
We selected GPT-4o to annotate the entire dataset of therapist utterances. Given its low mean entropy (0.035), we only ran the model once for full annotation. 

\begin{table}[t]
\centering
\resizebox{\linewidth}{!}{%
\small
\begin{tabular}{@{}ccccccc@{}}
\toprule
\textbf{Model} & \multicolumn{3}{c}{\textbf{No context}} & \multicolumn{3}{c}{\textbf{With context}} \\
\cmidrule(lr){2-4} \cmidrule(lr){5-7}
 & \textbf{FAC} & \textbf{PS} & \textbf{Overall} & \textbf{FAC} & \textbf{PS} & \textbf{Overall} \\
\midrule
GPT-4 & 0.65 & 0.76 & 0.68 &\textbf{0.65} & 0.70 & 0.64 \\
GPT-4o & \textbf{0.78} & \textbf{0.81} & \textbf{0.76} & 0.59 & \textbf{0.76} & \textbf{0.66} \\
Llama-3.1 & 0.66 & 0.70 & 0.61 & 0.55 & 0.65 & 0.50 \\
Yi-1.5 & 0.64 & 0.59 & 0.56 & 0.56 & 0.54 & 0.49 \\
\bottomrule
\end{tabular}}%
\caption{Average weighted F1 scores for PS Core (PS), Facilitators (FAC), and overall across different LLMs. The overall score is the weighted average of all strategies' F1 score plus None category.}
\label{tab:f1_scores}
\vspace{-0.35cm}
\end{table}

\begin{table}[t]
\centering
\resizebox{0.75\columnwidth}{!}{%
\small
\begin{tabular}{ccc}
\toprule
 \textbf{Model}  &  \textbf{Mean Entropy} &  \textbf{Std Entropy} \\
\midrule
 GPT-4  &         0.042 &        0.150 \\
GPT-4o  &         0.035 &        0.130 \\
 Llama  &         0.016 &        0.091 \\
 Yi-1.5 &         0.000 &         0.000 \\
\bottomrule
\end{tabular}}
\caption{Comparison of Mean and Standard Deviation of Entropy across different LLMs.}
\label{tab:entropy_comparison}
\vspace{-0.5cm}
\end{table}

\begin{table}[t]
\centering
\resizebox{\columnwidth}{!}{%
\small
\begin{tabular}{@{}clcccc@{}}
\toprule
&\textbf{Strategies} & \textbf{Precision} & \textbf{Recall} & \textbf{f1-score} & \textbf{\#Support} \\
\midrule
\multicolumn{1}{c}{\multirow{5}{*}{PS Core}}&Problem-solving Positive Mindset & 0.85 & 0.85 & 0.85 & 39 \\
&Defining Problems and Goals & 0.86 & 0.73 & 0.79 & 66 \\
&Generating Alternative Solutions & 0.83 & 0.76 & 0.79 & 36 \\
&Outcome Prediction and Planning & 0.86 & 0.82 & 0.84 & 33 \\
&Trying Out Solution Plan & 0.67 & 1 & 0.80 & 18 \\
\midrule
\multicolumn{1}{c}{\multirow{5}{*}{Facilitators}}&Social Courtesies & 0.59 & 0.94 & 0.73 & 25 \\
&Session Management & 0.58 & 0.81 & 0.68 & 64 \\
&Therapeutic Engagement & 0.69 & 0.95 & 0.80 & 130 \\
&Test Review & 1 & 0.86 & 0.92 & 24 \\
\midrule
&None & 0.88 & 0.48 & 0.62 & 90 \\
\midrule
&\textbf{weighted average} & 0.77 & 0.79 & 0.76 & 525 \\
\bottomrule
\end{tabular}}%
\caption{Precision, Recall, and F1-Score for Facilitative and PS Strategies in GPT-4o (without context)}
\label{tab:gpt4o_scores}
\vspace{-0.3cm}
\end{table}

\subsection{Transformer-Based Models}
Table \ref{tab:f1_scores_finetune} shows the average weighted F1 scores for DeBERTa, MentalBERT, FLAN-T5, and ModernBERT on PS Core and Facilitative strategies. ModernBERT achieved the highest F1 score (0.8) for PS Core strategies, while DeBERTa performed best on Facilitators (0.73). These results highlight DeBERTa's strength in Facilitators annotation and ModernBERT's advantage in PS Core (See Appendix \ref{Appendix_finetuned_results}).

\begin{table}[t]
\centering
\resizebox{0.6\columnwidth}{!}{%
\begin{tabular}{@{}lccc@{}}
\toprule
\textbf{Model} & \textbf{PS} & \textbf{FAC} & \textbf{Overall} \\ 
\midrule
DeBERTa    & 0.77  & \textbf{0.73}  & 0.68  \\ 
MentalBERT & 0.78  & 0.69  & 0.68  \\ 
ModernBERT& \textbf{0.8} & 0.7& \textbf{0.71} \\
FLAN-T5    & 0.68   & 0.69  & 0.67  \\ 
\bottomrule
\end{tabular}}
\caption{Average weighted F1 scores for PS Core (PS), Facilitators (FAC), and Overall across models.}
\label{tab:f1_scores_finetune}
\vspace{-0.5cm}
\end{table}

\begin{table*}[ht]
\centering
\resizebox{0.85\textwidth}{!}{%
\small
\begin{tabular}
{@{}cp{2.5cm}p{12cm}p{1.5cm}p{1.5cm}@{}}

\toprule
&\textbf{Strategies} & \textbf{Top Bigrams} & \textbf{Top LIWC} & \textbf{Distribution} \\
\midrule
&Defining Problems and Goals & (goals, around),   
(defining, problem),  
(facts, around),  
(obstacles, goals),  
(clear, language),  
(meeting, goals),  
(solving, important),  
(one, goals) &
reward
&33.48\%(1,743) \\
&Generating Alternative Solutions & (brainstorming, ideas), 
(strategies, tactics),  
(defer, judgment),  
(leads, quality),  
(quantity, leads),  
(three, things),  
(variety, ideas),  
(lots, ideas),  
(good, thing),  
(big, idea) &
insight, curiosity
&20.03\%(1,043) \\
\multicolumn{1}{c}{\multirow{5}{*}{\rotatebox{90}{PS Core}}}&Problem-solving Positive Mindset &  
(strong, emotion),  
(way, feel),  
(affects, way),  
(healthy, thinking),  
(make, mistakes),  
(positive, attitude),  
(things, change),  
(thinking, rules),  
(really, helpful),  
(use, feelings),  
(try, balance),  
(problems, challenges),  
(feelings, adaptively) &
emotion, feeling
&11.75\%(612) \\
&Outcome Prediction and Planning &  
(social, consequences),  
(personal, consequences),  
(positive, negative),  
(solve, problem),  
(pros, cons),  
(last, time),  
(personal, social),  
(rough, screening),  
(would, affect) &
risk, focusfuture
&10.18\%(530) \\
&Trying Out Solution Plan &  
(positive, negative),  
(see, works),  
(make, sense),  
(really, hard),  
(go, back),  
(solve, problem),  
(really, important),  
(effort, put),  
(get, support),  
(extra, support),  
(talk, trying) &
focuspast, focusfuture
&2.97\%(155) \\
\midrule
&Therapeutic Engagement & 
(really, good),  
(makes, sense),  
(right, right),  
(last, time),  
(oh, gosh),  
(good, good),  
(make, sure),  
(might, look),  
(one, things),  
(goals, around) & 
focuspresent
&52.19\%(5,918) \\
\multicolumn{1}{c}{\multirow{4}{*}{\rotatebox{90}{Facilitators}}}&Session Management & (next, time),  
(time, meet),  
(last, time),  
(defining, problem),  
(alternative, solutions),  
(next, week),  
(let, see),  
(going, talk),  
(time, talk),  
(brainstorming, ideas) &
focusfuture
&15.80\%(1,792) \\
&Social Courtesies & 
(thank, much),  
(look, forward),  
(thank, time),  
(talk, next),  
(taking, time),  
(next, time),  
(good, good),  
(thank, appreciate),  
(take, care),  
(really, helpful) &
emotion, prosocial
&8.69\%(986) \\
&Test Review & (problem, solving),  
(positive, rational),  
(impulsive, avoidant),  
(positive, negative),  
(high, positive),  
(rational, impulsive),  
(negative, orientation),  
(negative, rational),  
(approach, problem),  
(one, questionnaires),  
(positive, orientation),  
(five, dimensions) &
insight
&1.72\%(196) \\
\midrule
&None &-& - & 21.57\%(3,111) \\
\hline
\end{tabular}}
\caption{Overview of Strategies and Their Most Frequent Bigrams and LIWC features with Distribution}
\label{tab:strategies_dist}
\vspace{-0.3cm}
\end{table*}

\begin{figure*}
    \centering
    \begin{minipage}{0.49\textwidth}\includegraphics[width=0.9\columnwidth]{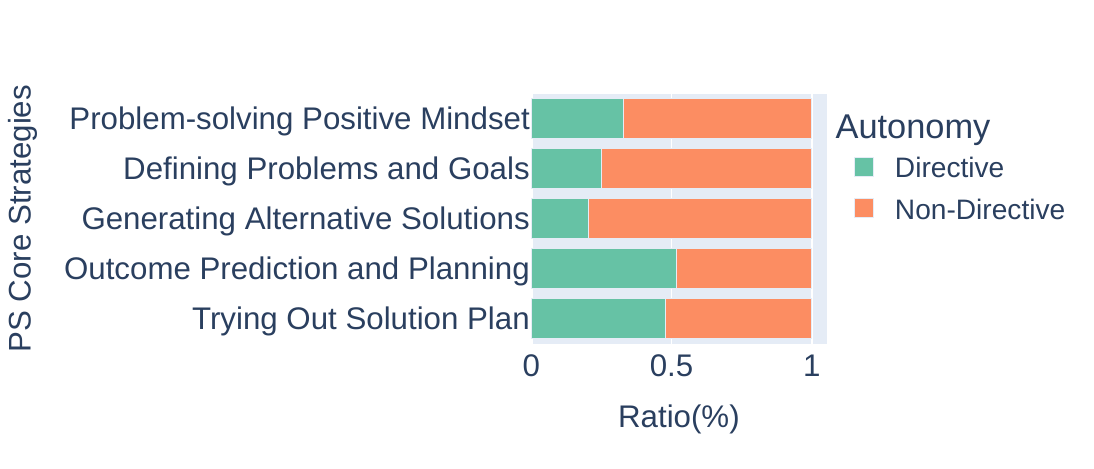}
    \vspace{-0.3cm}
    \caption{Distribution of Autonomy in PS Core}
    \label{fig:pscore_autonomy}
\end{minipage}%
\hfill
\begin{minipage}{0.49\textwidth}
    \centering
    \includegraphics[width=0.9\columnwidth]{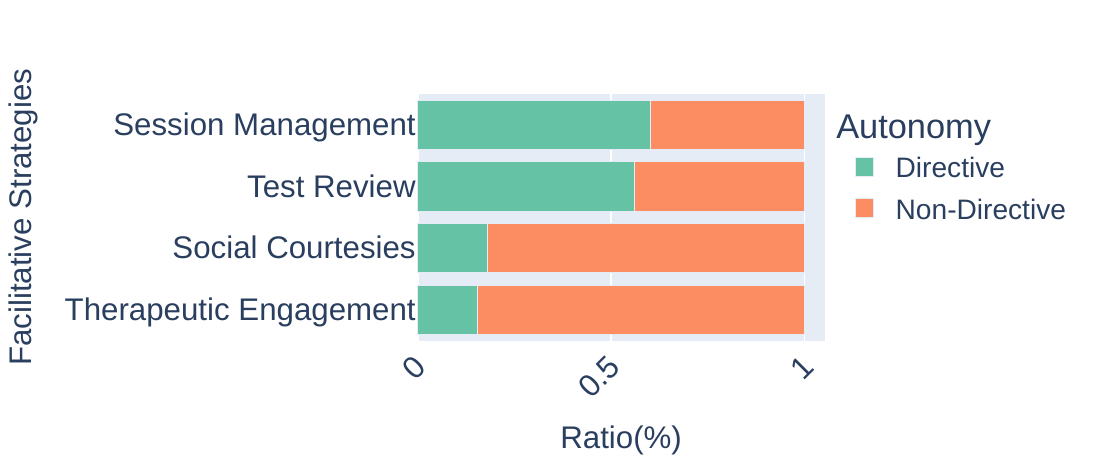}
    \vspace{-0.3cm}
    \caption{Distribution of Autonomy in Facilitators}
    \label{fig:fac_autonomy}
\end{minipage}%
\vspace{-0.5cm}
\end{figure*}

\subsection{Therapeutic Dynamics}
For autonomy, the relationship between directive and non-directive language and question types shows that 94.49\% of directive utterances are closed-ended. while 59.25\% of non-directive utterances are open-ended, a notable 40.75\% are unexpectedly closed-ended (See the co-occurrence matrix in Appendix \ref{Qtype_autonomy}). Figure \ref{fig:pscore_autonomy} shows the distribution of directive and non-directive autonomy in PS Core strategies. Non-directive language dominates the early stages (`PS Positive Mindset', `Defining Problems and Goals'), while the later stages (`Outcome Prediction and Planning', `Trying Out Solution Plan') shift toward directive language. Similarly, for Facilitators (Figure \ref{fig:fac_autonomy}) more structured strategies are directive while the other two show more non-directive language. 

\begin{figure*}
    \centering
    \begin{minipage}{0.49\textwidth}\includegraphics[width=0.9\columnwidth]{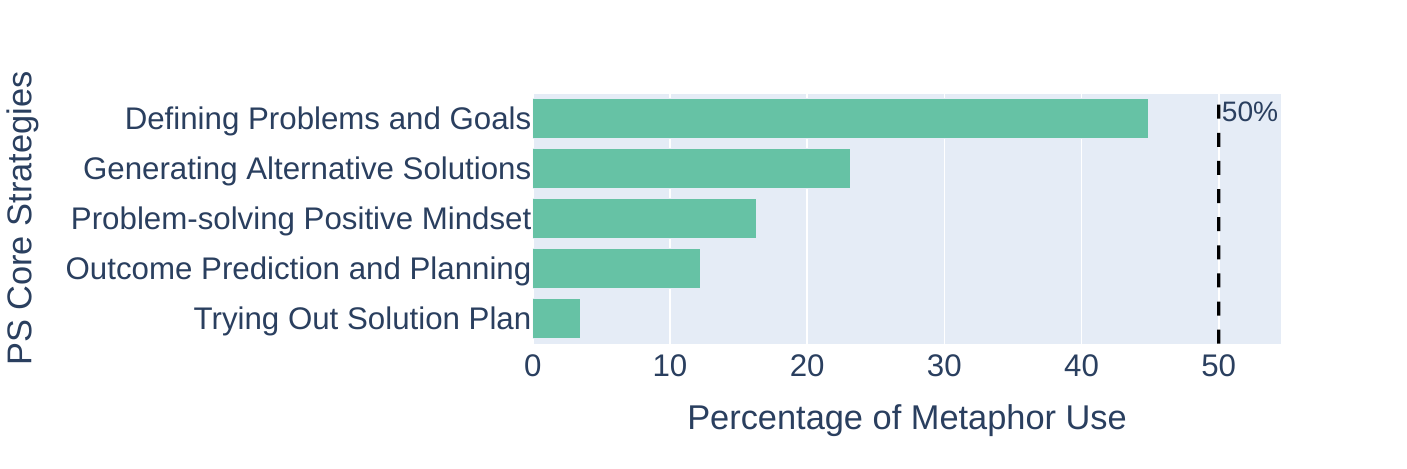}
    \caption{Proportion of Metaphors in PS Core}
    \label{fig:metaphor-pscore}
\end{minipage}%
\hfill
\begin{minipage}{0.49\textwidth}
    \centering
    \includegraphics[width=0.9\columnwidth]{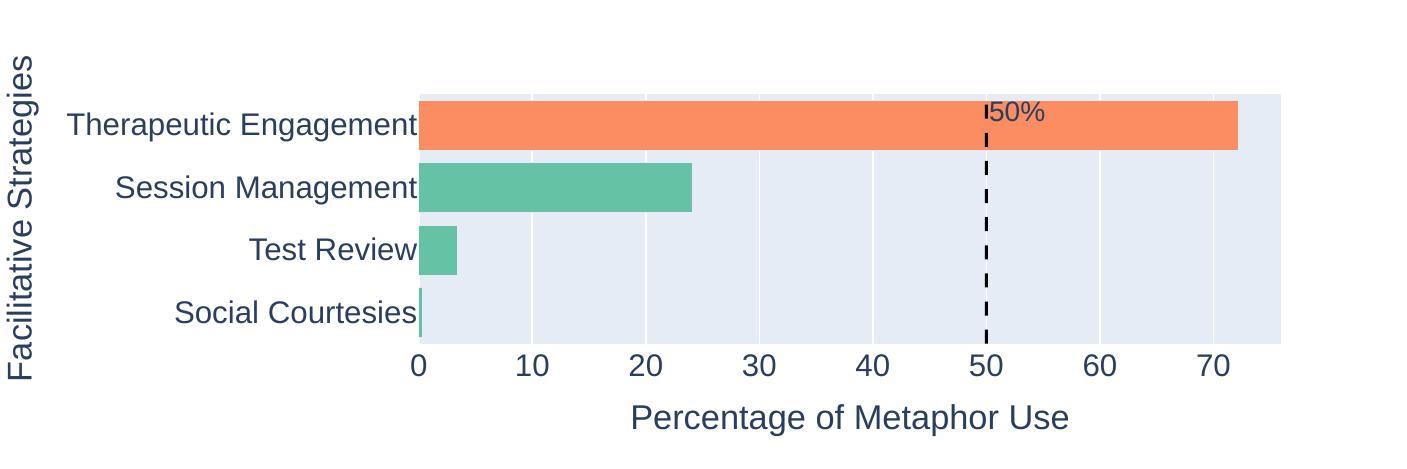}
    \caption{Proportion of Metaphors in Facilitators}
    \label{fig:metaphor-fac}
\end{minipage}%
\vspace{-0.5cm}
\end{figure*}

Reviewing the results of self-disclosure showed that, only 257 instances were labeled as non-immediate self-disclosure, suggesting that such disclosures are carefully considered rather than spontaneous. 
Furthermore, Figures \ref{fig:metaphor-pscore} and \ref{fig:metaphor-fac} show metaphor usage across PS Core and Facilitative Strategies. Metaphors are most prevalent in Therapeutic Engagement (72.18\%). 
In PS Core Strategies, they are commonly used in `Defining Problems and Goals' (44.90\%) and `Generating Alternative Solutions' (23.13\%).

\subsection{Linguistic Patterns}
Table \ref{tab:strategies_dist} shows the distribution of all therapeutic strategies in our data with the top bigrams, and LIWC categories representing each. 
Among PS Core strategies, `Defining Problems and Goals' is most prevalent (33.48\%), emphasizing therapists' role in helping clients clarify objectives, 
with bigrams like ``defining problem'' and ``solving important''. LIWC's `reward' feature suggests clients are being motivated by potential benefits. `Generating Alternative Solutions' is another significant strategy (20.03\%), with words such as ``brainstorming ideas'' and ``strategies tactics'' showing the focus on exploring different options to address issues. The LIWC indicators `insight' and `curiosity' align with this focus on creativity and open-mindedness.
Similarly, `Problem-Solving Positive Mindset' includes ``positive attitude'' and ``healthy thinking'', highlighting therapists' encouragement toward optimism and resilience, supported by LIWC features `emotion' and `feeling'. Outcome Prediction and Planning uses ``pros cons'' and ``solve problem'', emphasizing consequence evaluation. The prevalence of LIWC's `risk' and `focus future' reinforces a forward-thinking approach in decision-making.

On the `Facilitators' side, the bigrams highlight how therapists build rapport and guide session flow. In `Therapeutic Engagement', phrases like ``really good'' use affirming language to foster a supportive, collaborative atmosphere, helping clients feel understood. 
`Social Courtesies,' with phrases like `thank much' and `look forward,' add warmth to interactions, maintaining a respectful and friendly tone. The top LIWC features, `emotion' and `prosocial,' reflect the positive social behaviors therapists use to foster mutual respect and connection. Meanwhile, in `Test Review', more formal words like ``positive rational'' and ``impulsive avoidant'' suggest an analytical approach, where therapists review the client’s PST test results in a structured, evaluative manner (See Appendix \ref{liwc} for more information).

\subsection{Strategic Progression in Therapy Sessions.}
\begin{figure*}
    \centering
    \begin{minipage}{0.45\textwidth}\includegraphics[width=0.8\columnwidth]{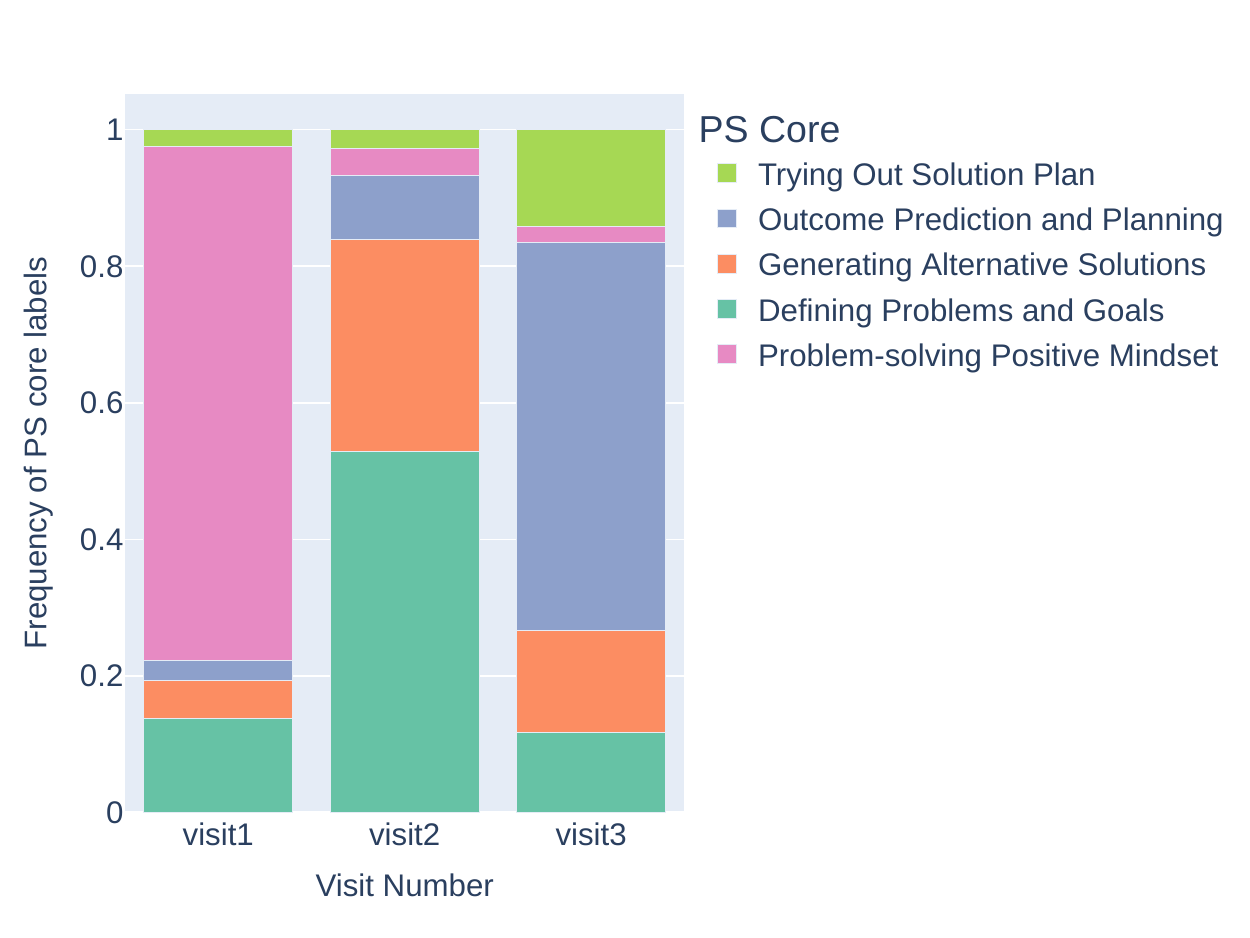}
    \caption{Distribution of PS Core in three visits}
    \label{fig:core}
\end{minipage}%
\begin{minipage}{0.45\textwidth}
    \centering
    \includegraphics[width=0.8\columnwidth]{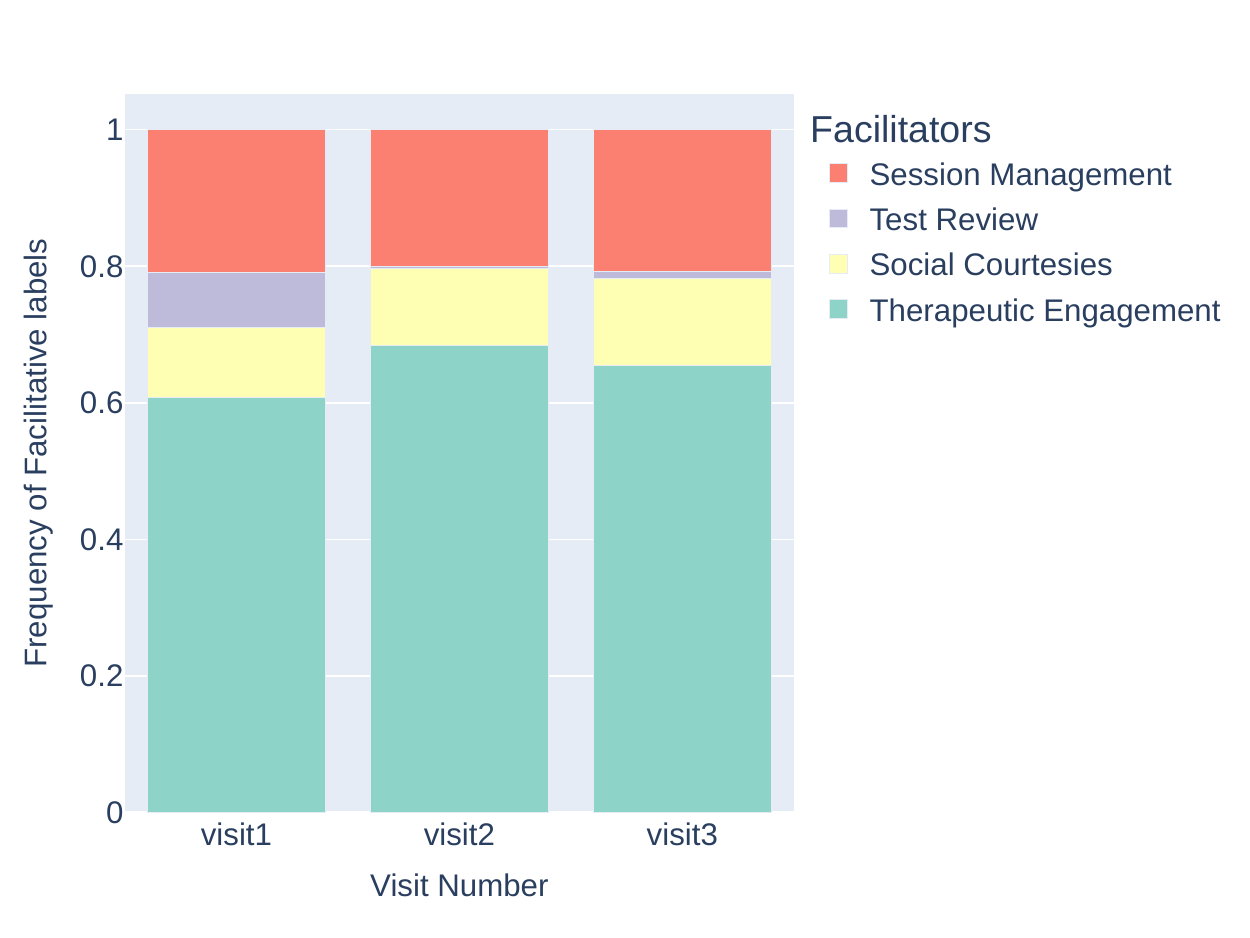}
    \caption{Distribution of Facilitators in three visits}
    \label{fig:comm}
\end{minipage}%
\vspace{-0.5cm}
\end{figure*}

We analyzed how strategies progress across three different visits of caregivers in our dataset. In Figure \ref{fig:core}, early sessions focus on enhancing `Problem-Solving Mindset' (Step One) and `Defining Problems and Goals' (Step Two), indicating that the initial phase of therapy is centered on improving caregivers' mindsets and helping them define the challenges they face. As therapy progresses into the second and third visits, there is a shift towards more advanced strategies such as `Generating Alternative Solutions' (Step Three) and `Outcome Prediction and Planning' (Step Four), showing that the focus moves towards exploring possible solutions and planning for their implementation. In the final phase (Visit Three), `Trying Out Solution Plan' (Step Five) becomes prominent, indicating a transition to applying and testing solutions in real life.
Figure \ref{fig:comm} shows the consistent use of Facilitative strategies like `Therapeutic Engagement', `Session Management', and `Social Courtesies', reinforcing their essential role throughout therapy. However, `Test Review' is more frequent in Visit One, reflecting the therapist's focus on reviewing the client's Test results taken prior to starting the session before progressing to intervention. This early emphasis lays the groundwork for later sessions. These patterns further validate the effectiveness of our classifier.

\section{Discussion} \label{sec:discussion}

\noindent\textbf{Model Performance and the Need for Open-Source Improvements.} 
Our results show that GPT-4o outperformed GPT-4 and open-weight models in annotating therapeutic strategies, achieving an F1 score of 0.76 without supplemental context. While Yi had the lowest entropy, GPT-4o showed greater consistency and reliability, highlighting the trade-off between performance stability and model reliability in mental health applications. Despite GPT-4o's success, improving open-source models is crucial for enhancing accuracy, accessibility, and competition in healthcare AI. From an ethical perspective, open-source models promote transparency and accountability, ensuring that healthcare AI advancements benefit a broader community and expand access to high-quality mental health resources—unlike proprietary models, which can limit access and obscure critical decision-making processes \cite{welsh2024democratising}.

\noindent\textbf{Improving LLMs for Emotional and Social Understanding.}
GPT-4o excelled in labeling structured, goal-oriented PS core strategies, achieving high precision and recall. This highlights LLMs' strength in processing clear, task-oriented language. However, GPT-4o struggled with Facilitative strategies such as `Session Management' and `Social Courtesies', which rely on emotional and social cues. This underscores a key limitation: while LLMs handle problem-solving tasks well, they struggle with the nuances of interpersonal communication, lacking the ability to interpret human emotions accurately. These challenges arise from implicit, context-dependent interactions, which LLMs often misinterpret, sometimes displaying an over-realistic sense of empathy \cite{sorin2023large}. To improve effectiveness in therapeutic settings, future work should refine LLMs using expert-in-the-loop systems, allowing for real-time human feedback to enhance emotional and social cue recognition \cite{afzal-etal-2024-towards, 10.1145/3581754.3584142}.

\noindent\textbf{Efficient and Privacy-Conscious Therapy Annotation. } Our fine-tuned transformer-based models show the effectiveness of the distillation (student-teacher) method, achieving strong performance. ModernBERT outperformed our second-best LLM, GPT-4, with an overall F1 score of 0.71. It achieved 0.8 F1 in PS Core strategies, while DeBERTa specialized in Facilitative strategies, reaching an F1 score of 0.73. This suggests that smaller models can effectively annotate distinct therapeutic strategies, offering a scalable and efficient alternative for therapy content analysis. Additionally, these models provide a privacy-conscious solution for processing sensitive therapy data, ensuring confidentiality and trust in digital therapeutic environments \cite{10.1145/3555640}.

\noindent\textbf{How PST is Delivered. }Beyond classifying strategies, understanding how therapists apply therapeutic tools like autonomy, self-disclosure, and metaphor offers deeper insight into PST's communication dynamics. In early PS Core stages, therapists favor non-directive language to foster client exploration, shifting to directive language later for clarity and structure, aligning with a client-centered problem-solving approach \cite{nezu2012problem}. Examining question types and autonomy, we found that some closed-ended questions still encouraged elaboration, which our domain expert termed ``imposters.'' For example, ``Do you have any questions about what we've gone over or for this process?''—though closed-ended, it invites further discussion, blurring the line between directive and non-directive communication. This suggests that therapists sometimes frame questions strategically to encourage deeper engagement, even when using traditionally restrictive formats.

Therapists also use non-immediate self-disclosure strategically and selectively, as reflected in the low frequency of such utterances in our dataset. In Facilitators, self-disclosures often involve personal anecdotes to build rapport, whereas, in PS Core, they serve a more functional role, modeling problem-solving strategies for clients. This finding is supported by previous work \cite{knox2003therapist} which identified similar patterns of strategic disclosure timing and emphasized how self-disclosure can serve as different functions at different stages of therapy.
Metaphors help illustrate ideas and enhance emotional connection in Therapeutic Engagement \cite{lyddon2001metaphor, tay2014metaphor}, while in PS Core strategies, they help conceptualization and creativity. However, they decline in action-oriented phases, where direct communication becomes essential. Therapists adapt their language to session goals—fostering exploration early on and shifting to structured, goal-driven communication as therapy progresses. 

\noindent\textbf{Broader Implications. }
Our model has meaningful real-world applications for PST practice and documentation. It can assist annotating and analyzing therapy sessions, helping document which PST strategies were employed and how effectively they were implemented, and serving valuable educational purposes for therapists-in-training.
As LLM-assisted therapy models become more prevalent, our work provides important guidance for human-AI collaborative approaches. By identifying which PST strategies are most crucial and how they should be properly sequenced for building client rapport, we can better inform the development of LLM-assisted therapy systems where human therapists remain in the loop. 
The model can also serve as an evaluation framework for assessing whether AI-assisted therapy is appropriately implementing PST strategies and maintaining therapeutic alliances through their interactions.



\vspace{-0.25cm}
\section{Conclusion} \label{sec:conclusion}\vspace{-0.25cm}
This study highlights the potential of LLMs like GPT-4o for effective annotation of therapeutic dialogues. GPT-4o achieved a notable F1 score of 0.76, with a low entropy value of 0.035, demonstrating its efficiency and reliability in categorizing PST strategies. We fine-tuned transformer-based models on a sample of 5,000 therapist utterances annotated by GPT-4o and showed that ModernBERT exceeded the performance of our second-best LLM, GPT-4, showcasing the effectiveness of using LLM-generated annotations to fine-tune smaller, domain-specific models.
We also expanded PST codebook to include a new Facilitators dimension, enriching our analysis of therapist-client interactions. 
We also analyzed Therapeutic Dynamics ( autonomy, self-disclosure, and metaphor) and linguistic patterns in our data to better understand `how' PST strategies are used and in what contexts. 
Our research offers valuable insights into enhancing technology integration in mental health care, aiming to make therapeutic interventions both more accessible and more effective.
\section{Limitations} \label{sec:limitation}
Our study has several limitations, including the dependence on `transcribed' therapy sessions, which may not fully capture the dynamic aspects of live interactions such as tone and non-verbal cues. The use of LLMs like GPT-4o and GPT-4 introduces potential biases from their internet-based training data, impacting the interpretation of therapeutic strategies. Furthermore, our analysis is solely based on English-language data, which limits the model's applicability in non-English-speaking contexts. These limitations highlight the technology's current stage and its focus on textual analysis, excluding the broader emotional and psychological facets of therapy sessions.

In addition, we only focus on PST-related framework and analysis which may not be generalizable to other therapeutic methods like cognitive-behavioral therapy (CBT) or interpersonal therapy (IPT). These methods may involve different communication styles and dynamics that the LLMs we used may not accurately interpret. Future work will analyze the adaptability of our model for broader therapy analysis.
Additionally, cultural and contextual differences in therapy could further limit the model's applicability across diverse mental health settings. Future research could address this by testing the models in a broader range of therapeutic approaches and contexts.

Finally, we did not compare our results with any LLM-generated therapy data, which could offer a valuable comparison between real-world therapeutic interactions and simulated data produced by LLMs. Our focus is exclusively on the characteristics of authentic, real-world data from therapy sessions. Future research could explore this comparison, examining how well LLM-generated interactions align with or diverge from real-world therapeutic PST dialogues, potentially offering insights into improving the practicality and effectiveness of AI-driven interventions in mental health care.

\section{Ethical Considerations} \label{sec:ethics}
We are collaborating with a team of PST experts who were in charge of collecting and transcribing the data. We only have access to secondary data, and the primary data was collected as part of a larger task approved by the primary institution's review board. As a result, we do not have access to any identifiable information about the clients or therapists. We understand that using anonymized therapy data raises concerns regarding privacy and the potential for re-identification of people involved in these sessions. We ensured that all data used in our analysis was thoroughly cleaned, with the help of our collaborators, and that strict measures were put in place to remove any remaining information from the transcriptions. 

The application of LLMs in healthcare, particularly in sensitive areas like therapy, necessitates careful consideration of their implications. Ensuring responsible development and use of AI technologies in mental health care is crucial, requiring vigilance for biases, stakeholder engagement, and strong safeguards to maintain patient confidentiality and the integrity of the therapeutic process. 
This paper does not endorse the use of LLMs in therapeutic settings, nor does it claim that they are ready for such applications. Rather, our goal is to systematically evaluate the behavior of current LLMs when applied in therapeutic contexts.
The use of AI in mental health care raises ethical concerns about automating therapy. AI tools, like LLMs, should assist rather than replace therapists, ensuring they do not undermine the personal, empathetic nature of therapy, which is crucial for effective treatment. While AI can provide useful insights, therapists must remain the primary decision-makers, as AI cannot fully grasp the complexities of human behavior and emotions as shown in our results and other related work. Accountability should always rest with human professionals to preserve the integrity and effectiveness of mental health care.
\section{Acknowledgment} \label{sec:acknowledgment}
We are deeply grateful to Dr. Bum Chul Kwon, Max Song, and DongWhan Jun for their invaluable insights and assistance throughout this project. This paper is based upon work supported by the National Science Foundation under Grant \#2144880.

\bibliography{custom, colm2024_conference}

\appendix

\section{Appendix}\label{sec:appendix}

\section{Therapists Strategies}
Table \ref{codebook} presents the final list of therapist strategies for PST, with definitions and examples for each category. 

\begin{table*}[t]
\centering
\resizebox{0.9\textwidth}{!}{%
\small
\begin{tabular}{@{}cp{0.2\textwidth}p{0.39\textwidth}p{0.32\textwidth}@{}}
   \toprule
Category&Strategy & Definition & Example \\ \midrule
&Problem-solving Positive Mindset (Step One) & Strategies building a positive approach towards challenges, emphasizing optimism and emotional awareness. & \textit{``uh, they suggest trying to think of problems as challenges rather than threats.''} \\ 
&Defining Problems and Goals (Step Two) & Setting achievable goals and identifying factual problems while recognizing obstacles. & \textit{``Are there any obstacles to those goals? Can we list anything that is getting in the way?''} \\ 
\multicolumn{1}{c}{\multirow{5}{*}{\rotatebox{90}{PS Core}}}&Generating Alternative Solutions (Step Three) & Brainstorming solutions, differentiating between strategies, and encouraging open idea exploration. & \textit{``So let's brainstorm some specific strategies and tactics for how you can make that happen.''} \\ 
&Outcome Prediction and Planning (Step Four) & Narrowing down solutions, assessing impacts, and creating an action plan. & \textit{``Working through this worksheet...short-term personal consequences of self-care are pretty positive, right?''} \\ 
&Trying Out Solution Plan (Step Five) & Implementing and assessing the solution strategy, troubleshooting, and acknowledging successes. & \textit{``Then just monitoring your outcomes. Has the problem improved in some ways?''} \\ 
\midrule
&Social Courtesies* &  Polite, surface-level interactions like greetings, small talk, and farewells to create a friendly atmosphere. & \textit{``thank you again for your time today, and have a really good week.''} \\ 
\multicolumn{1}{c}{\multirow{4}{*}{\rotatebox{90}{Facilitators}}}&Session Management* & The therapist provides guidance on session logistics, manages time, keeps discussions on track, and helps plan future sessions. & \textit{``So you got Skype up and running, and that is so fun!''} \\
&Therapeutic Engagement* & The therapist shows attention, validates emotions, offers support, reflects feelings, praises strengths, and provides comfort or personal insight. & \textit{``I’m glad it's starting to show some small improvements.''} \\
&Test Review* & The therapist explains the PS Test and reviews the client's results across five dimensions: positive/negative problem orientation, rational, impulsive, and avoidant problem-solving styles. & \textit{``So positive and rational were your highest scores.  Your positive was, uh, really, is almost as high as it could possibly get.''} \\
\bottomrule
\end{tabular}}
    \caption{Overview of Therapist Strategies for PST, *denotes newly created codes}
    \label{codebook}
\end{table*}

\section{Model Setting} \label{app:model}
\paragraph{LLM parameters: }
\texttt{temperature = 0;}\\
\texttt{max-tokens = 500;}\\
\texttt{openai-api-version=`gpt-4o-2024-05-13',  `gpt-4-0125-preview', `meta-llama/Llama-3.1-70B-Instruct', `01-ai/Yi-1.5-34B-Chat'}
\paragraph{Transformer-based Models Parameters: } Adam optimizer, \texttt{learning-rate=2e-5, }
\texttt{DeBERTa = microsoft/deberta-v3-base, }
\texttt{MentalBERT = mental/mental-bert-base-uncased, }
\texttt{FlanT5 = google/flan-t5-base} 
\texttt{ModernBERT = answerdotai/ModernBERT-base}

\section{Prompts} \label{Appendix_prompts}
In this section, we present the specific prompt used to obtain annotations from LLMs (the few-shot examples are not included here).

\subsection{Classification of PST Strategies}
\subsubsection{Without Context}

\paragraph{Instruction.} \texttt{You are a highly helpful assistant tasked with annotating the therapist's behavior. Your role is to categorize the therapist's behavior using a list of strategies provided below.}

\texttt{Your Task:
Select one strategy from the `PS core' category and one strategy from the Facilitators category that best aligns with the therapist's utterance.
If none of the strategies are applicable, return 'None' for that category.}

\paragraph{Definition. } \texttt{Definitions of Therapist's Strategies for Problem-Solving Therapy:}

  \texttt{PS core Strategies:
    1- Problem-solving Positive Mindset: Techniques that foster a positive outlook, such as promoting optimism, reframing negative thoughts, visualizing success, practicing gratitude, mindfulness, positive self-talk, and emotional awareness.
    2- Defining Problems and Goals: Identifying clear, achievable goals, recognizing obstacles, and determining which specific issues to address in therapy.
    3- Generating Alternative Solutions: Brainstorming a variety of solutions without judgment, distinguishing between strategies and tactics, and encouraging open exploration of ideas.
    4- Outcome Prediction and Planning: Narrowing down solutions, evaluating their impact (personal, social, temporal), and creating a detailed action plan for implementation.
    5- Trying Out Solution Plan: Implementing the solution, assessing its effectiveness, troubleshooting issues, seeking support, and acknowledging successes in the process.}

    \texttt{Facilitative Strategies:
    6-  Social Courtesies: Engaging in polite, yet somewhat superficial social interactions like greetings, small talk (e.g., asking about the weather or weekend plans), thanking the client for their time, and saying goodbye.
    7- Session Management: Managing session logistics, such as informing the client about session length, guiding the direction of the conversation, refocusing when off-topic, or scheduling future sessions.
    8- Therapeutic Engagement: Actively listening, validating the client's feelings, providing emotional support, recalling previous information shared by the client, reflecting emotions, normalizing experiences, commenting on client strengths, offering personal disclosure, and offering comfort.
    9- Test Review: Discussing the client's Problem-Solving Test results, explaining the test, and reviewing five problem-solving dimensions: positive problem orientation, negative problem orientation, rational style, impulsive style, and avoidant style.}
\subsubsection{With Context}
\paragraph{Instruction. } \texttt{You are a highly helpful assistant tasked with annotating the therapist's behavior. Your role is to categorize the therapist's behavior using a list of strategies provided below.}

\texttt{Your Task:
Select one strategy from the PS Core category and one strategy from the Facilitators category that best aligns with the therapist's most recent utterance. You are also provided with the last two utterances: the therapist's previous utterance and the client's previous utterance. You should use only these two previous utterances as context, but for annotation, you should focus only on the most recent therapist's utterance. If none of the strategies are applicable, return 'None' for that category.}

\paragraph{Definition. }The definition is the same as when context is not provided.
\subsection{Extracting Domain Specific Features} \label{Appendix_prompts_domain}
\paragraph{Prompt. } 
\texttt{You are a helpful assistant tasked with analyzing a therapist's utterances in a problem-solving therapy session. Your goal is to assess the utterances for the following aspects:}
      \texttt{1. Autonomy-Supportive Language
      Output Labels: "Directive", "Non-Directive", or "N/A".
      Definitions:
    - Directive:
      - Focuses on the therapist’s goals, plans, or strategies for the client.
      - Therapist explicitly tells the client what to do or prescribes a solution.
      - Prioritizes achieving a specific outcome, often determined by the therapist.
      - May reflect the therapist’s beliefs, goals, or values rather than the client’s.
      - Frequently uses closed-ended or leading questions to guide the client toward a particular answer or decision.
      - Provides answers, advice, or logical consequences rather than encouraging exploration.
    - Non-Directive:
      - Focuses on the client’s goals, values, and intrinsic motivations.
      - Encourages the client to explore their thoughts and emotions freely without being led toward a specific conclusion.
      - Gives the client control over the direction of therapy and supports their decision-making process.
      - Avoiding judgment or imposition of therapist’s opinions.
      - Uses open-ended questions to encourage self-reflection and insight.
      - Attends to the client’s feelings and underlying experiences, fostering understanding and autonomy.
      - Accepts and explores resistance as part of the process, fostering collaboration to address challenges.}

      \texttt{2. Self-Disclosure
      Output Labels: "Immediate", "Nonimmediate", or "N/A".
      Definitions: Self-disclosure refers to instances when the therapist intentionaly shares personal information, experiences, or feelings with the client.
      - Immediate Self-Disclosure: Sharing personal thoughts, feelings, or reactions related to the ongoing therapeutic interaction.
      - Nonimmediate Self-Disclosure: Discussing past experiences, personal anecdotes, or biographical information unrelated to the immediate session.}
      
      \texttt{3. Open vs. Closed Questions
      Output Labels: "Open-Ended", "Closed-Ended", or "N/A".
      Definitions:
      - Open-Ended Questions: Questions that encourage detailed, thoughtful, and expansive responses. These questions often begin with "how," "what," "why," or "tell me about" and invite the client to explore their thoughts and feelings freely.
      - Closed-Ended Questions: Questions that elicit specific, often short responses such as "yes," "no," or concise information. These questions are structured and limit the scope of the client’s answer.
      N/A: Use this label if the utterance does not fit into either category.}

      \texttt{4. Metaphors:
      Output Labels: "yes, <reasoning for metaphor detection>" (for use of metaphor) or "no" (for no use of metaphor).
s      Definition: According to conceptual metaphor theory, metaphor facilitates a mapping of attributes or characteristics from source domain to target domain. For example, the word “invested” in the sentence “I have
      invested a lot of time in her” is a metaphorical expression. The source domain implied by this metaphor is the domain of money and the target domain implied by this metaphor is the domain of time. Based on the above information, decide whether this utterance uses metaphor in it and mention the phrase, source and domain of the metaphor.}

      Required Output Format (JSON):
                    
                        "autonomy": "<string>",
                        "self disclosure": <integer>,
                        "question type": "<string>",
                        "metaphor": "<string>"

\section{Detailed Results of Classification}
\subsection{LLM Classification Results} \label{Appendix_llm_results}
The detailed classification results for each strategy, based on the F1-scores of different LLMs, are presented in Table \ref{tab:llms_f1_scores}.

\begin{table*}[t]
\centering
\resizebox{0.85\textwidth}{!}{%
\begin{tabular}{@{}clcccc@{}}
\toprule
& \textbf{Strategies} & \textbf{GPT-4} & \textbf{GPT-4o} & \textbf{Llama} & \textbf{Yi} \\
\hline
\multicolumn{1}{c}{\multirow{5}{*}{PS Core}} & Problem-solving Positive Mindset        & 0.83 & \textbf{0.85} & 0.76 & 0.75 \\
& Defining Problems and Goals             & 0.73 & \textbf{0.79} & 0.72 & 0.63 \\
& Generating Alternative Solutions        & 0.74 & \textbf{0.79} & 0.62 & 0.48 \\
& Outcome Prediction and Planning         & 0.74 & \textbf{0.84} & 0.67 & 0.50 \\
& Trying Out Solution Plan                & \textbf{0.89} & 0.80 & 0.75 & 0.40 \\
\hline
\multicolumn{1}{c}{\multirow{5}{*}{Facilitators}}& Social Courtesies                       & \textbf{0.75} & 0.73 & 0.58 & 0.67 \\
& Session Management                      & \textbf{0.72} & 0.68 & 0.61 & 0.61 \\
& Therapeutic Engagement                  & 0.57 & \textbf{0.80} & 0.69 & 0.64 \\
& Test Review             & 0.89 & \textbf{0.92} & 0.79 & 0.70 \\
\hline
& None                                    & 0.61 & \textbf{0.62} & 0.28 & 0.27 \\
\hline
& weighted avg                            & 0.68 & \textbf{0.76} & 0.61 & 0.56 \\
\bottomrule
\end{tabular}}
\caption{F1-scores of different LLMs for each strategy (without context)}
\label{tab:llms_f1_scores}
\end{table*}

\subsection{Fine-tuned Models Classification Results} \label{Appendix_finetuned_results}
The detailed classification results for each strategy, based on the F1-scores of different fine-tined transformer-based models, are available in Table \ref{tab:finetuned_f1_scores}.

\begin{table*}[t]
\centering
\resizebox{0.85\textwidth}{!}{%
\begin{tabular}{@{}clccc@{}}
\toprule
& \textbf{Strategies} & \textbf{DeBERTa} & \textbf{MentalBERT} & \textbf{FlanT5} \\
\midrule
\multicolumn{1}{c}{\multirow{5}{*}{PS Core}} & Problem-solving Positive Mindset        & \textbf{0.81} & 0.78 & 0.76 \\
& Defining Problems and Goals             & \textbf{0.80} & 0.79 & 0.79 \\
& Generating Alternative Solutions        & 0.73 & 0.70 & \textbf{0.75} \\
& Outcome Prediction and Planning         & 0.73 & \textbf{0.77} & 0.70 \\
& Trying Out Solution Plan                & 0.80 & \textbf{0.89} & 0.40 \\
\hline
\multicolumn{1}{c}{\multirow{5}{*}{Facilitators}} & Social Courtesies                       & \textbf{0.73} & 0.67 & 0.68 \\
& Session Management                      & \textbf{0.66} & 0.63 & 0.62 \\
& Therapeutic Engagement                  & \textbf{0.70} & 0.68 & 0.69 \\
& Test Review             & \textbf{0.83}& 0.79 & 0.80 \\
\hline
& None                                    & 0.41 & \textbf{0.53} & 0.50 \\
\hline
& weighted avg                            & \textbf{0.68} & \textbf{0.68} & 0.67 \\
\bottomrule
\end{tabular}}
\caption{F1-scores of different fine-tuned models for each strategy}
\label{tab:finetuned_f1_scores}
\end{table*}

\section{LIWC Results} \label{liwc}
\begin{figure*}
    \centering
    \includegraphics[width=0.8\textwidth]{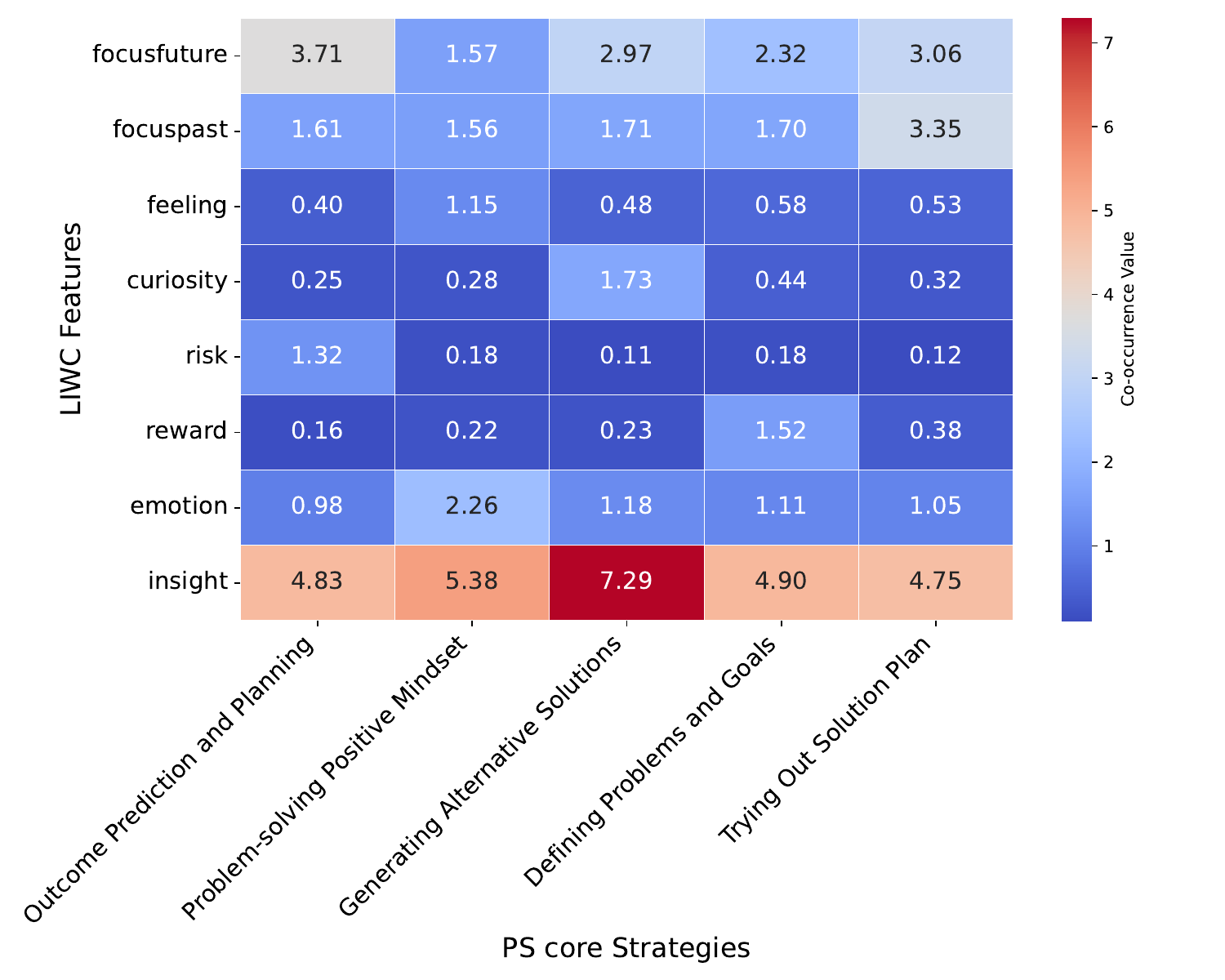} 
    \caption{Co-Occurrence of PS core Strategies with Selected LIWC Features}
    \label{fig:liwc_core}
\end{figure*}

\begin{figure*}
    \centering
    \includegraphics[width=0.8\textwidth]{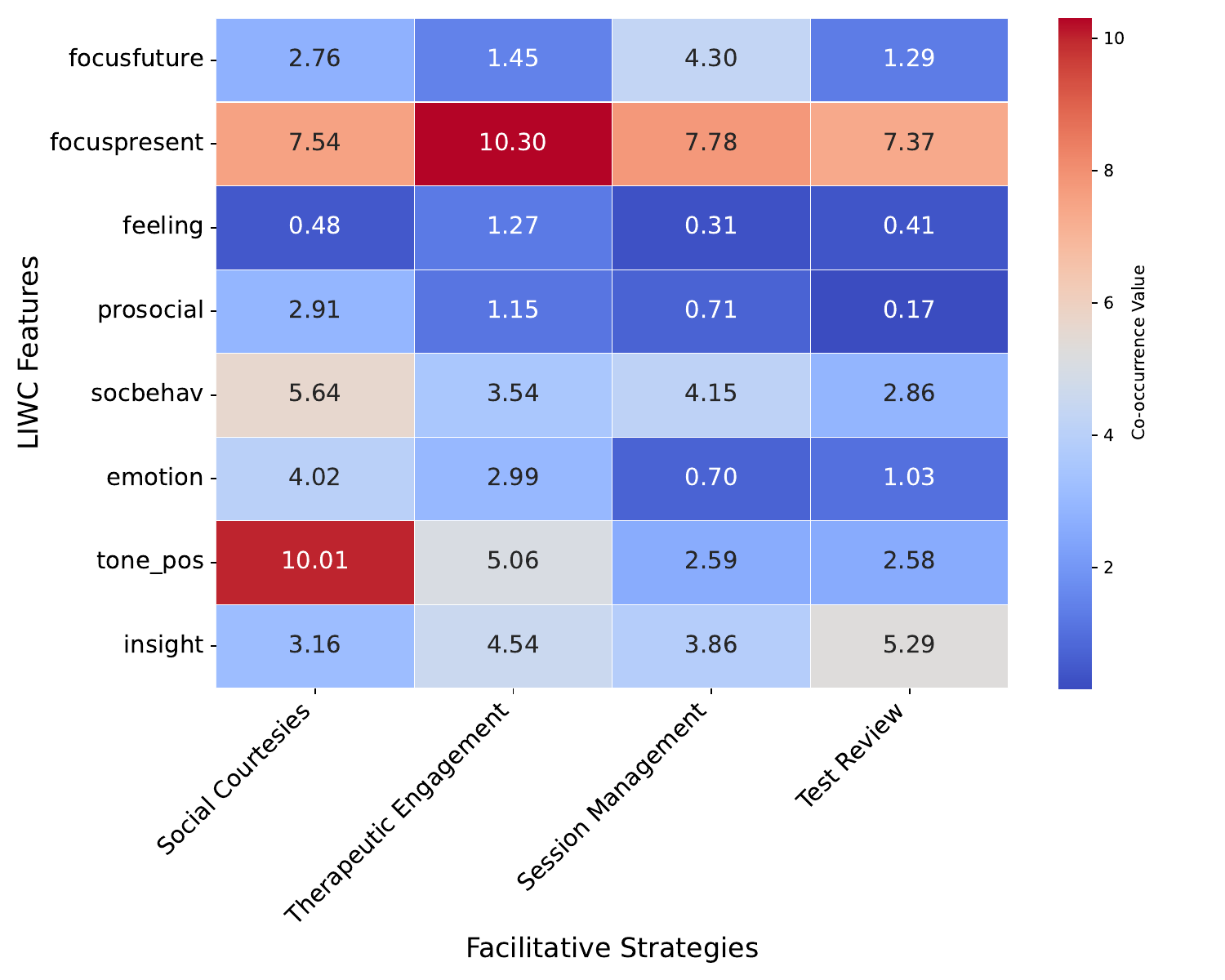} 
    \caption{Co-Occurrence of Facilitative Strategies with Selected LIWC Features}
    \label{fig:liwc_comm}
\end{figure*}

Figures \ref{fig:liwc_core} and \ref{fig:liwc_comm} present the co-occurrence of each strategy with the selected LIWC features.

\section{Question Type vs. Autonomy} \label{Qtype_autonomy}
\begin{figure}
    \centering
    \includegraphics[width=.99\linewidth]{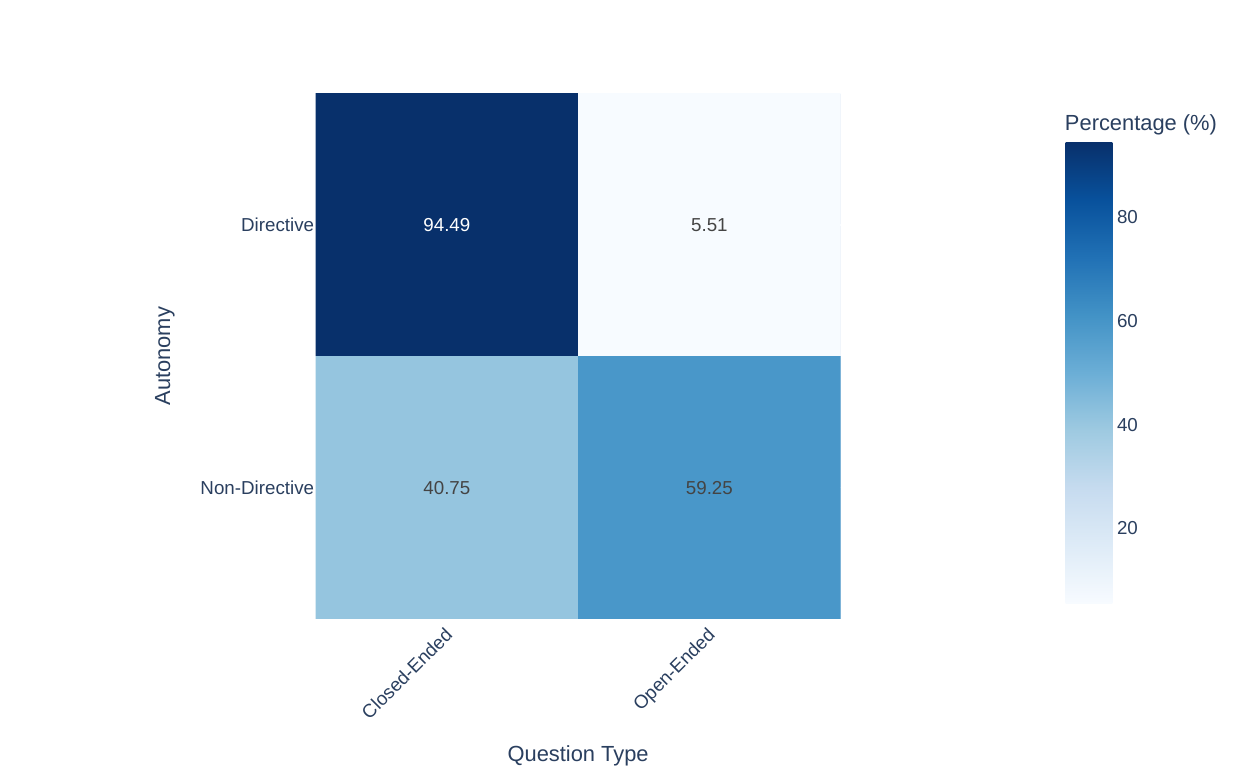}
    \caption{Co-occurrance of Question Type and Autonomy Labels}
    \label{fig:Qtype_autonomy}
    \vspace{-0.5cm}
\end{figure}

Figure \ref{fig:Qtype_autonomy} shows the co-occurrence matrix between question type and autonomy labels.

\end{document}